\documentclass[12pt, a4paper, notitlepage, oneside]{article}

\usepackage{booktabs}
\usepackage{graphicx}
\usepackage[utf8]{inputenc}
\usepackage{amsmath}
\usepackage{amssymb}
\usepackage[hidelinks]{hyperref}
\usepackage{algorithm}
\usepackage{algpseudocode}
\usepackage{refcount}
\usepackage{rotating}
\usepackage{caption}
\usepackage{subcaption}
\usepackage{geometry}
\newgeometry{vmargin={25mm,50mm}, hmargin={35mm,35mm}}
\usepackage{fancyhdr}
\fancypagestyle{plain}{
\fancyhf{}
\pagestyle{fancy}
\fancyhead[C]{\includegraphics[height=3cm]{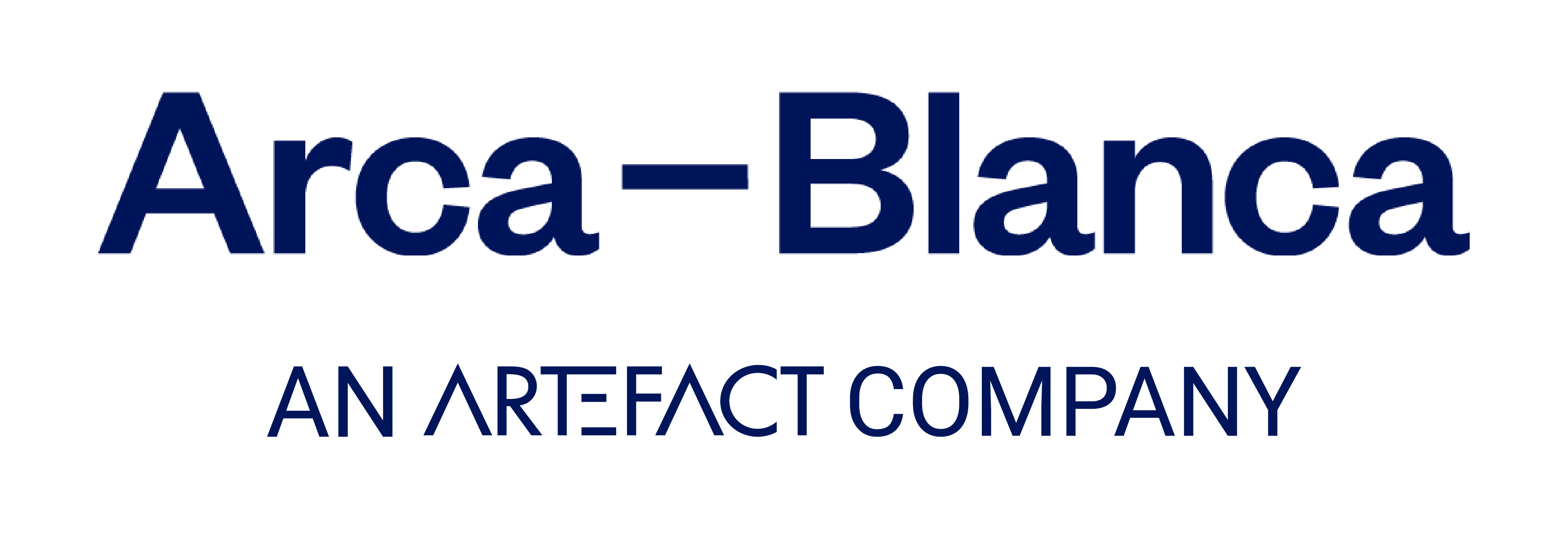}}
\fancyfoot[L]{Copyright \textcopyright \vspace{6pt} 2021 Arca Blanca}
\fancyfoot[R]{ \thepage}
}
\fancypagestyle{AB}{
\fancyhf{}
\pagestyle{fancy}
\fancyhead[L]{\includegraphics[height=3cm]{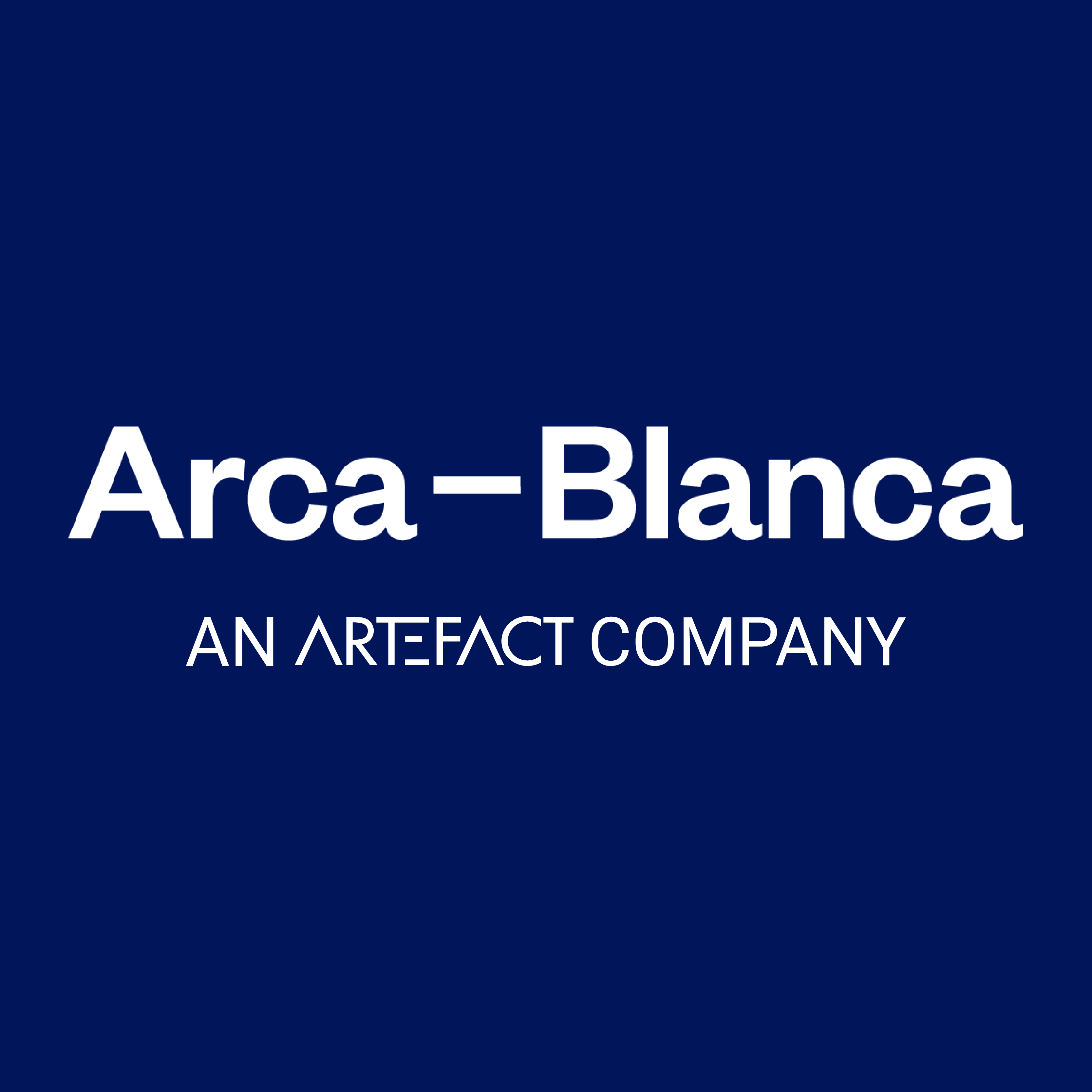}}
\fancyfoot[L]{Copyright \textcopyright \vspace{6pt} 2022 Arca Blanca}
\fancyfoot[R]{ \thepage}
}
\usepackage{authblk}
\title{Aggregating Macroeconomic Forecasts}
\author{  
  Hurwitz, Evan\\
  \texttt{evan.hurwitz@arcablanca.com}
  \and
  Peace, Nelson\\
  \texttt{nelson.peace@arcablanca.com}
   \and
  \v{C}evora, George\\
  \texttt{george.cevora@arcablanca.com}
  }
\affil{Arca Blanca Ltd.\\Manfield House, One Southampton St, London WC2R 0LR, United Kingdom}
\affil{Artefact Research Centre, 19 Rue Richer, 75009 Paris, France}
\date{}
\newcommand\fref[1]{%
    (\getrefnumber{#1})\ifcase\getrefnumber{#1}\fi}

\title{ Achieving Stable Training of Reinforcement Learning Agents in Bimodal Environments through Batch Learning}

\date{}

\begin{document}

\maketitle

\begin{abstract}
Bimodal, stochastic environments present a challenge to typical Reinforcement Learning problems. This problem is one that is surprisingly common in real world applications, being particularly applicable to pricing problems. In this paper we present a novel learning approach to the tabular Q-learning algorithm, tailored to tackling these specific challenges by using batch updates. A simulation of pricing problem is used as a testbed to compare a typically updated agent with a batch learning agent. The batch learning agents are shown to be both more effective than the typically-trained agents, and to be more resilient to the fluctuations in a large stochastic environment. This work has a significant potential to enable practical, industrial deployment of Reinforcement Learning in the context of pricing and others.
\\

\end{abstract}

\begin{figure}
    \centering
    \includegraphics[width=10cm]{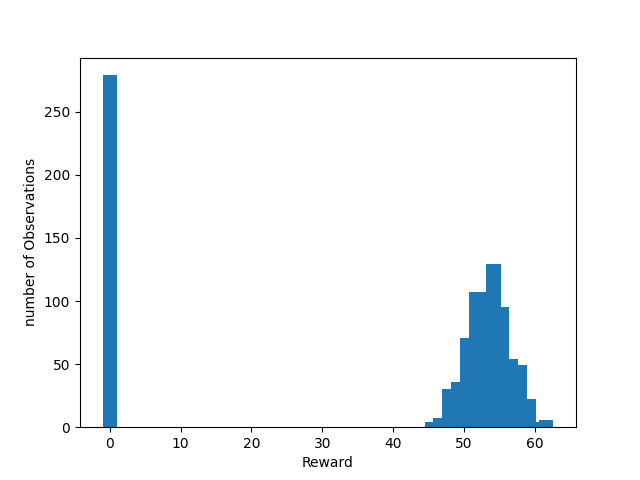}
    \caption{Bimodal reward signals resulting from a theoretical pricing problem. Some customers do not purchase and produce reward signal of 0\$, others purcahse at a varying level of discount.}
    \label{fig:Bimodal}
\end{figure}

Much of the work and literature within the field of Reinforcement Learning [RL] focuses on problem domains with well-regulated rewards \cite{regulated-rewards} \cite{regulated-rewards2}. These rewards fall into expected ranges with similar orders of magnitude. This ignores a common problem of failure states, in which a very different reward, typically zero, would be received that is very dissimilar to the reward received in an episode resulting in a success state \cite{dissimilar-rewards}. This results in a bimodal distribution of rewards, as shown in Figure \fref{fig:Bimodal}. While typical methods of applying RL will have some degree of success, the bimodal nature of the reward distribution disrupts the learning capability of the agent considerably. This paper explores a novel batch-learning approach to the problem in order to circumvent this disrupted learning pattern, smoothing out the learning process and thus facilitating better performance.

A commercial pricing problem is an apt example to support our discussion: decision makers are commonly presented with difficult decisions surrounding what discount if any to their products will result in optimal revenue generation \cite{discount-problem-commonality}. The bimodal distribution of rewards emerges as a result of binary consumer behaviors regardless of what discount is applied: consumers will either purchase or not. The resulting reward distribution therefore has two modes 1) for the purchases at varying discount levels, 2) zero when the customer doesn't purchase. Using conventional RL techniques this results in an unstable and unpredictable training process.

Our investigation was performed in an experimental manner, by first creating a parameterised model environment that produces the problematic behaviour and then creating agents that use RL to solve the environment across its parameterisations, both with and without the proposed batch learning innovation. These results are then contrasted in order to arrive at robust conclusions as to the effectiveness of the batch learning innovation and the conditions under which it is most impactful. 

\section{Methods}
In line with the commercial pricing example we construct a toy-problem to demonstrate the efficiency of batch learning. The experiments will allow an agent offering discounts and learning in a RL fashion to interact with simulated environment. A number of variants on the environment and agent will be explored.

\subsection{Environment}
Pricing problems are a simple yet relevant context to investigate bimodal rewards  by virtue of only one decision stage occurring in the environment; a purchase or not. The environment is set up such that it contains a customer who has a base probability of purchasing a given product, and that the customer's probability of purchase will increase or decrease in a nonlinear manner dependant on the discount offered. 

The bimodal distribution captures the probability of no purchase, or failure, corresponding to a return of \$0, while also providing approximately normally distributed values, representing the returns in the instance of a purchase, or success as shown in Figure \ref{fig:Bimodal}. Two distinct problems are captured in this broader RL problem domain. The first problem is to determine whether the result will be a success or a failure. The second problem is to map the degree of success if success occurs. 
These two problems can result in significant challenges for a conventional RL solution, as the failure modes skew the q-values recorded in the q-table, particularly in instances where early training results in an outlier volume of sequential failures.

The agent interacts with the environment through offering discounts $d$ from the base price $\pi$ to individual customers thus providing reward $R=\pi(1-d)$ to the agent. 
Individual customers $c$ vary in their base probability of considering a purchase at all $P(\beta_c)$ but if they do consider a purchase, the probability of purchase $\gamma$ increases as a function of discount $d$ according to $P(\gamma|\beta_c)=1-e^{\zeta d}$ where $\zeta$ is an arbitrary constant.
This results in an overall probability of purchase $\tau$ for a given customer of $P{\tau_c}=P{\beta_c}-P(\neg \gamma| \beta_c, d)$ and the expected reward from offering a discount to customer $\mathbb{E}(R_c)=P({\tau_c})R(\pi,d)$. 
The essential quantities are summarised in figure \ref{fig:quantities} for the sake of clarity. 

\begin{figure}[!h]
\centering
\begin{tabular}{l|p{8cm}|l}
& meaning & properties \\
\hline
$c$ & customer identity & \\
$\pi$ & base (undiscounted) price &  $ R \in \mathcal{R^+}$\\
$d$ & discount & $d \in \mathcal{R}:\langle0,1\rangle$ \\
$R$ & reward / revenue, objective of maximisation through RL &  $ R \in \mathcal{R^+}$\\
$\beta$ & considering a purchase at all irrespective of discount & binary event\\
$\gamma$ & purchase event, given a discount $d$ and an initial consideration $\beta$ & binary event\\
$\tau$ & purchase & binary event \\
\end{tabular}
\caption{The fundamental quantities describing the environment of our RL task.}
\label{fig:quantities}
\end{figure}

\begin{figure}[!h]
    \centering
    \includegraphics[width=10cm]{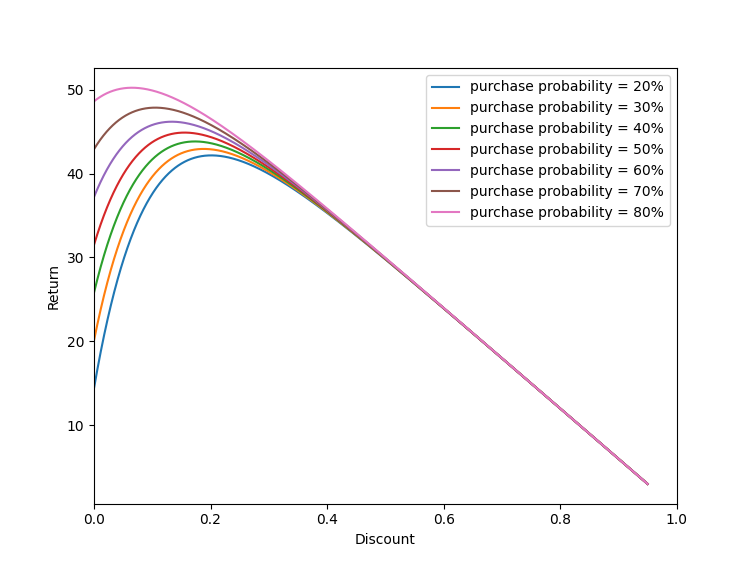}
    \caption{Expected revenue $\mathbb{E}(R_c|d,\beta)$ as a function of discount and different probabilities of initial consideration of the purchase by customers.}
    \label{fig:exp_returns}
\end{figure}

While $P(\gamma)$ increases as the discount increases, the revenue resulting from the purchase decreases with the discount increasing. This results in a non-monotonic relationship between discount and expected revenue $\mathbb{E}(R_c|d)$ shown in Figure  \fref{fig:exp_returns}. The optimum discount is therefore the value which maximises $\mathbb{E}(R_c)$. This value can be found in an analytical fashion for the environment defined here, but is unknown in the real world. The task of RL agents tested in this environment is to maximise the function without knowledge of the environment. 

While the $\mathbb{E}(R_c|d,\beta)$ function shown in Figure \ref{fig:exp_returns} is smooth, and with no local minima, the optimisation task is not trivial due to the bimodal nature of the rewards. To explore the characteristics of the task we set up two different versions of the environment:
\begin{itemize}
    \item The initial consideration of purchase $\beta$ varying in probability between 0.2 and 0.8 in 0.1 steps, called the sparse state-space. Resulting in $P(\beta)\in[0.2; 0.3; 0.4; 0.5; 0.6; 0.7; 0.8]$.
\item The initial consideration of purchase $\beta$ varying in probability between 0.2 and 0.785 in 0.015 steps, called the granular state-space. Resulting in 40 different values of $P(\beta)$.
\end{itemize}

\subsection{Agent}
The agent is required to make a single decision what discount to offer the customer at a single time-point. To accommodate a standard, tabular q-learning approach, this action-space is discretised. Two versions of the action-space discretisation are explored, each representing a different level of granularity:

\begin{itemize}
    \item The agent may choose from 10 discount levels, the sparse action-space: $d\in[0; 0.1; 0.2; 0.3; 0.4; 0.5; 0.6; 0.7; 0.8; 0.9]$.
    \item The agent may choose from 81 discount levels in 0.01 intervals between 0 and 0.79, the granular action-space.
\end{itemize}

The agent learns using standard tabular Q-learning with Bellman equation \cite{Bellman-equation}, making decisions according to an an $\epsilon$-greedy policy \cite{e-greedy-policy}. The learning rate used is 0.1, and the $\epsilon = 0.9$. 

\subsection{Experimental setup}
To demonstrate the benefit of batch learning for the task defined here the agent will either update the Q-table after each observation as is common in online RL or alternatively store a user-defined number (1000 in the experiments presented here) of observations in memory before updating the Q-table with average rewards over the whole batch. Factorial design is used to explore all four combinations of state- and action- spaces.

All experiments are run for 100,000 iterations allowing for learning until convergence. Convergence is defined as the agent achieving (on a rolling mean basis) 95\% of the optimum performance under perfect knowledge of the environment. The optimum was calculated analytically for sparse state-space, but approximated using MCMC for the granular state-space.

\subsection{Illustrative results from a simplified experiment}

In order to demonstrate the shortfalls and merits of single-update and batch-update visually to the reader an additional experiment is run in a more simplistic environment. For this purpose the base purchase probability is fixed at 0.6 making 0.1 discount the optimum action for the agent. 

\begin{figure}[!h]
     \centering
     \begin{subfigure}[b]{0.49\textwidth}
         \centering
         \includegraphics[trim={0 0 0 1.4cm},clip,width=\textwidth]{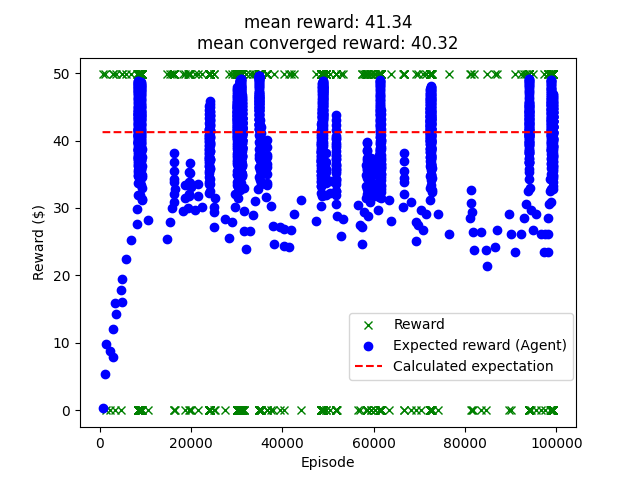}
         \caption{Standard Q-learning}
         \label{fig:learningDynamics_standard}
     \end{subfigure}
     \begin{subfigure}[b]{0.49\textwidth}
         \centering
         \includegraphics[trim={0 0 0 1.4cm},clip,width=\textwidth]{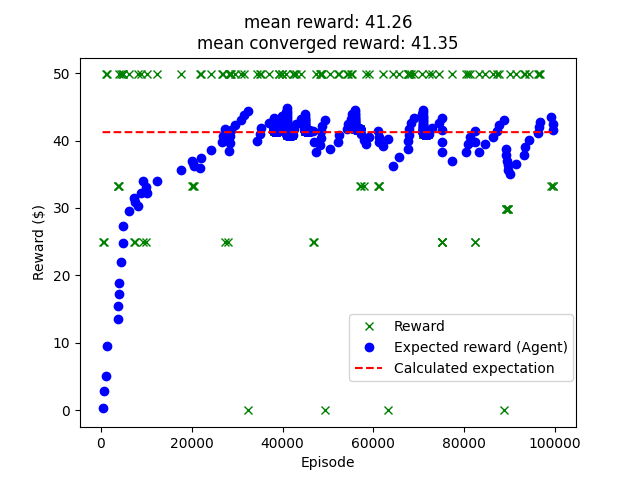}
         \caption{Batch Q-learning}
         \label{fig:learningDynamics_batch}
     \end{subfigure}
        \caption{Learning dynamics in a simplified experimental setup to facilitate understanding of agent's performance. The agent in both traditional Q-learning and in the batched Q-learning cases have the full range of 80 actions to choose from, and their Q-states and rewards are tracked at each step. The illustrative results are then shown for a single action $d = 0.17$, with events from all other actions filtered out. This allows us to observe the change in Q-value expectations as well as the reward signals received by the agent in sequential order.}
        \label{fig:learningDynamics}
\end{figure}

As can be seen in Figure \ref{fig:learningDynamics} both standard and batch Q-learning results in a gradual convergence towards the analytically calculated expectation, which is inaccessible to the agent, then oscillate around that value in response to the stochasticity of the rewards. Compared to the standard approach, batch learning offers slower convergence, but is less sensitive to the stochastic oscillations, with higher overall reward achieved, making it much more stable and usable tool to be deployed. Convergence is clearly slower, but this is more than made up for in better converged performance which comes out to a 2.17\% overall improvement in reward over the time period, as well as more stable converged performance.

\section{Results}
Four unique experimental setups shown in figure \ref{table:optimal rewards} were tested in eight experiments, with batch learning outperforming the standard approach in every single case as measured by the total reward accumulated over the experiments. The final reward at fully-trained state was improved by batch learning in all cases except for the largest state-space.

\begin{figure}[H]
\centering
\begin{tabular}{p{0.1666\textwidth}p{0.1466\textwidth}p{0.10666\textwidth}p{0.14666\textwidth}p{0.1866\textwidth}}
Action-space size & State-space size & Update method& Final \nobreak{reward} [\$]& Total \linebreak reward [$10^{6}$\$]\\
\hline
10 & 10 & single & 38.18 &  3.35   \\
10 & 10 & batch & 38.73 &  3.46 \\
10 & 40 & single & 37.77 &  3.31 \\
10 & 40 & batch  & 38.58 &  3.35 \\
80 & 10 & single & 38.66 &  3.31 \\
80 & 10 & batch & 38.93 &  3.42 \\
80 & 40 & single & 37.97 &  3.11 \\
80 & 40 & batch & 37.41 &  3.17 \\
\end{tabular}

\caption{Benchmark rewards vs RL agent learned rewards for both q-learning methodologies}
\label{table:optimal rewards}
\end{figure}

\section{Conclusions}

This paper proposes a batch training method for Reinforcement Learning [RL] agents. This update method has been shown to be superior compared to standard methods in environments with bimodal rewards such as offering discounts to customers with differing price sensitivity. The batch training method not only produces better performance in these challenging environments, but also produces more decisive agents acting on learned information. Convergence time is adversely affected, as is intuitive, which becomes the cost for a more stable and optimal converged state when operating in a bimodal manner.
Stable performance and better convergence in real-world scenarios like this is an important step on the path to industrialize  RL.

\bibliography{bib}
\bibliographystyle{plain}
\end{document}